\documentclass{article}
\usepackage[accepted]{eiml_icml2026}

\usepackage{hyperref}
\usepackage{amsmath,amssymb,amsthm}
\usepackage{booktabs}
\usepackage{graphicx}
\usepackage{enumitem}
\usepackage{caption}
\usepackage{multirow}
\usepackage{microtype}
\usepackage{bm}
\usepackage{float}
\usepackage{placeins}


\newcommand{\Sph}{\mathbb{S}^2}

\newcommand{\R}{\mathbb{R}}

\newcommand{\Prob}{\mathbb{P}}
\newcommand{\dgeo}{d_{\mathrm{geo}}}

\newcommand{\calC}{\mathcal{C}}
\newcommand{\calD}{\mathcal{D}}
\newcommand{\calM}{\mathcal{M}}
\newcommand{\calX}{\mathcal{X}}
\newcommand{\qhat}{\hat{q}}
\newcommand{\shat}{\hat{\sigma}}
\newcommand{\yhat}{\hat{y}}

\newcommand{\Bgeo}{B_{\mathrm{geo}}}

\icmltitlerunning{Geometry-Aware Conformal Prediction on Manifolds}
\begin{document}

\twocolumn[
\icmltitle{Geometry-Aware Uncertainty Quantification via\\Conformal Prediction on Manifolds}

\icmlsetsymbol{equal}{*}

\begin{icmlauthorlist}
\icmlauthor{Marzieh Amiri Shahbazi}{rit}
\icmlauthor{Ali Baheri}{rit}
\end{icmlauthorlist}

\icmlaffiliation{rit}{Rochester Institute of Technology, Rochester, NY, USA}

\icmlcorrespondingauthor{Marzieh Amiri Shahbazi}{ma7684@rit.edu}
\icmlcorrespondingauthor{Ali Baheri}{akbeme@rit.edu}

\icmlkeywords{Conformal Prediction, Riemannian Manifolds, Uncertainty Quantification, Geodesic Distances}

\vskip 0.15in
]

\printAffiliationsAndNotice{}

\begin{abstract}
Conformal prediction gives finite-sample coverage guarantees for regression, but most standard constructions are designed for Euclidean output spaces. When the response lies on a Riemannian manifold, Euclidean residuals and coordinate-based regions can ignore the geometry that defines meaningful error. We propose \emph{adaptive geodesic conformal prediction}, a simple framework that builds nonconformity scores from geodesic distances and normalizes them with a cross-validated estimate of local prediction difficulty. On the sphere, this produces geodesic caps whose area is independent of position, while their radii still adapt to heteroscedastic noise. In both a synthetic sphere experiment and an IGRF-14 geomagnetic field forecasting task, the adaptive method preserves valid marginal coverage, reduces variation in conditional coverage, and improves worst-case coverage relative to non-adaptive and coordinate-based baselines.
\end{abstract}

\section{Introduction}
\label{sec:intro}

Many scientific prediction problems have responses that are not naturally Euclidean. Geomagnetic field directions lie on the unit sphere \citep{alken2021igrf}, protein backbone dihedral angles live on the flat torus \citep{ramachandran1963stereochemistry}, spacecraft orientations belong to the rotation group \citep{peretroukhin2020smooth}, and brain connectivity patterns can be represented as symmetric positive definite matrices \citep{yuan2012local}. Point prediction for such data has matured over the last decade \citep{petersen2019frechet, lin2017extrinsic, fletcher2013geodesic}. The accompanying uncertainty quantification problem is less settled: \emph{how should one build reliable prediction sets when the response space itself is curved?}

Conformal prediction (CP) is attractive because it turns almost any predictive model into a set-valued predictor with finite-sample, distribution-free coverage under exchangeability \citep{vovk2005algorithmic, lei2018distribution}. The framework has also been extended to covariate shift \citep{tibshirani2019conformal}, improved conditional behavior \citep{romano2019conformalized}, and non-exchangeable data \citep{barber2023conformal}. However, a direct Euclidean implementation is often poorly matched to manifold-valued responses. Two issues are especially important.

\emph{First, the output geometry is part of the statistical problem.} Standard CP typically measures residuals with Euclidean or coordinate-based distances. On a curved space, these distances can depend on the chosen chart rather than on the intrinsic discrepancy between two responses. On the sphere, for example, a fixed-width coordinate rectangle has an area that changes with latitude, becoming very small near the poles and largest near the equator. A geodesic cap with the same intrinsic radius has the same area at every location.

\emph{Second, prediction difficulty is rarely constant.} In geomagnetic forecasting, secular variation is much faster in some regions of the globe than in others. A constant-width prediction set therefore tends to over-cover easy regions and under-cover hard regions. In Euclidean regression, a common remedy is to normalize residuals by a local difficulty estimate \citep{lei2018distribution, romano2019conformalized}. For manifold-valued responses, the same idea should be carried out using geodesic distances and a geometry-aware notion of residual size.

\textbf{Related work.}
CP on manifolds remains comparatively underexplored. \citet{kuleshov2018conformal} builds prediction regions in a learned embedding, while our goal is to construct regions directly on the output manifold. \citet{messoudi2021copula, messoudi2022ellipsoidal} study multi-target regression with copula-based and ellipsoidal regions in Euclidean space. Other recent work has broadened CP beyond the standard i.i.d.\ Euclidean setting: \citet{shahbazi2026hierarchical} combines group-aware conformal calibration with Bayesian posterior uncertainties for hierarchical healthcare data; \citet{baheri2025conformal} intersects scale-specific prediction sets for multi-resolution problems; and \citet{millard2025split} extends split conformal guarantees to infinite-dimensional function spaces with neural operators. These directions address important sources of structure in the data, but they do not explicitly handle the intrinsic geometry of the response space.

\textbf{Contributions.}
We develop \emph{adaptive geodesic conformal prediction} for manifold-valued regression. The method keeps the usual split-conformal guarantee while replacing coordinate-dependent residuals with intrinsic ones.
\begin{enumerate}[leftmargin=*,topsep=2pt,itemsep=1pt]
\item \textbf{Geometry-aware conformal regions.} We form prediction sets with geodesic distances, yielding regions that respect the manifold geometry, such as spherical caps with location-independent area.
\item \textbf{Locally adaptive calibration.} We scale residuals by a covariate-dependent difficulty estimate trained only on the proper training data, so the final regions expand in harder parts of the input space without using calibration labels for model fitting.
\item \textbf{Empirical validation.} On a synthetic sphere benchmark and an IGRF-based geomagnetic forecasting task, the method improves conditional coverage uniformity and worst-case coverage while maintaining valid marginal coverage.
\end{enumerate}

\section{Method}
\label{sec:method}

We study regression with covariates $X \in \calX \subseteq \R^p$ and responses $Y$ on a complete Riemannian manifold $(\calM, g)$ with geodesic distance $\dgeo$. For a target miscoverage level $\alpha$, we want a prediction region $\calC(X) \subseteq \calM$ such that $\Prob(Y_{n+1} \in \calC(X_{n+1})) \ge 1 - \alpha$. As usual in conformal prediction, this is a marginal coverage statement. Our practical objective is also to make coverage as stable as possible across values of $X$, while avoiding unnecessarily large regions.

The experiments focus on the unit sphere $\Sph = \{y \in \R^3 : \|y\| = 1\}$. On $\Sph$, the natural distance is $\dgeo(u,v) = \arccos(\langle u, v\rangle)$, and a geodesic ball is a spherical cap with area $2\pi(1-\cos r)$. This area depends only on the radius $r$, not on where the cap is centered.

\subsection{Conformal Prediction with Geodesic Scores}
\label{sec:conformal}

We use the split conformal framework \citep{vovk2005algorithmic, lei2018distribution}. The data are divided into a training set $\calD_{\mathrm{tr}}$, a calibration set $\calD_{\mathrm{cal}}$, and a test set $\calD_{\mathrm{te}}$. The training set is used to fit a point predictor $\yhat : \calX \to \calM$ and a difficulty estimator $\shat : \calX \to \R_+$. The calibration set is then used only to choose the final scale of the prediction regions. For each calibration point, we compute a nonconformity score $s(X_i, Y_i)$, take the conformal quantile $\qhat_\alpha$ at level $\lceil(1-\alpha)(n_{\mathrm{cal}}+1)\rceil / n_{\mathrm{cal}}$, and return $\calC(x) = \{y \in \calM : s(x,y) \le \qhat_\alpha\}$.

We compare three scores. The proposed \textbf{adaptive geodesic} score divides the geodesic residual by a local estimate of prediction difficulty,
\begin{equation}\label{eq:score_adaptive}
    s_{\mathrm{adapt}}(x, y) \;=\; \dgeo\bigl(\yhat(x),\, y\bigr) \;/\; \shat(x),
\end{equation}
This yields a geodesic ball $\calC(x) = \Bgeo(\yhat(x),\, \qhat_\alpha \cdot \shat(x))$. Its radius is smaller where the fitted model is locally reliable and larger where prediction is harder. The \textbf{standard geodesic} score uses the unnormalized distance $s_{\mathrm{geo}}(x,y) = \dgeo(\yhat(x), y)$, which gives a constant-radius ball. The \textbf{naive coordinate} score applies an $L^\infty$ distance in spherical coordinates $(\theta,\varphi)$ and therefore produces axis-aligned coordinate rectangles.

On $\Sph$, both geodesic scores produce spherical caps with position-independent area. In contrast, a coordinate rectangle has area proportional to $\sin\theta$ and can become intrinsically narrow near the poles. The adaptive score is the only score among the three that changes the region size from point to point, which is helpful when the noise level changes over the input space.

\textbf{Difficulty estimation.}
The difficulty estimator $\shat(x)$ must be trained without looking at the calibration labels. We use only $\calD_{\mathrm{tr}}$. First, we compute 5-fold cross-validated residuals, $e_i = \dgeo(\yhat^{(-k)}(X_i), Y_i)$. Second, we fit a $k$-NN regressor with $k_\sigma = 20$ to predict these residuals from the covariates, and we clip the prediction below at $\epsilon > 0$ for numerical stability.

\textbf{Conditional coverage evaluation.}
To assess whether coverage is uniform, we sort test points into six equal-frequency bins by estimated difficulty $\shat(x)$ and compute coverage in each bin. We summarize the result with the standard deviation of bin-level coverage and the worst bin coverage.

Algorithm~\ref{alg:main} summarizes the full workflow.

\begin{algorithm}[H]
\caption{Adaptive Geodesic Conformal Prediction}
\label{alg:main}
\footnotesize
\begin{algorithmic}[1]
\REQUIRE Data $\{(X_i, Y_i)\}_{i=1}^n$, manifold $(\calM, \dgeo)$, level $\alpha$
\ENSURE Prediction regions $\{\calC(X_i)\}$ with coverage $\ge 1-\alpha$
\STATE \textit{Training (uses $\calD_{\mathrm{tr}}$ only):}
\STATE Fit base predictor $\yhat$ on $\calD_{\mathrm{tr}}$
\STATE Compute CV residuals: $e_i = \dgeo(\yhat^{(-k)}(X_i), Y_i)$
\STATE Fit difficulty estimator $\shat$ by regressing $e_i$ on $X_i$
\STATE \textit{Calibration (uses $\calD_{\mathrm{cal}}$ only):}
\STATE Compute $s_i = \dgeo(\yhat(X_i), Y_i) / \max(\shat(X_i), \epsilon)$
\STATE $\qhat_\alpha \gets$ conformal quantile of $\{s_i\}$
\STATE \textit{Prediction:}
\FOR{each test point $x$}
\STATE $\calC(x) \gets \Bgeo\bigl(\yhat(x),\; \qhat_\alpha \cdot \max(\shat(x), \epsilon)\bigr)$
\ENDFOR
\end{algorithmic}
\end{algorithm}

\section{Riemannian Geometry Preliminaries}
\label{sec:geometry}

We briefly recall the spherical geometry used in the experiments. On the unit two-sphere~$\Sph$, the geodesic distance is the great-circle distance $\dgeo(u,v)=\arccos(\langle u,v\rangle)\in[0,\pi]$. A spherical cap of radius~$r$ has area $2\pi(1-\cos r)$, so the same intrinsic radius always corresponds to the same area.

In standard spherical coordinates $(\theta,\varphi)$ with colatitude $\theta\in[0,\pi]$ and azimuth $\varphi\in[-\pi,\pi)$, the metric is $\mathrm ds^2=\mathrm d\theta^2+\sin^2\!\theta\,\mathrm d\varphi^2$. An $L^\infty$-ball of half-width~$\delta$ in these coordinates defines a coordinate rectangle whose area is
\begin{equation}\label{eq:naive_area_s2}
    \mathrm{Area}_{\text{naive}}(\delta,\theta)
    =2\delta\,\bigl|\cos(\theta-\delta)-\cos(\theta+\delta)\bigr|.
\end{equation}
Unlike the cap area, this expression depends on~$\theta$. Near the poles, the azimuthal direction is metrically compressed, so a fixed-width coordinate rectangle becomes intrinsically too narrow and may under-cover. This is a direct consequence of representing a curved surface with coordinates that distort area.

\section{Experiments}
\label{sec:experiments}

We compare three conformal prediction methods on~$\Sph$:
\textbf{Adaptive Geodesic} (proposed), \textbf{Standard Geodesic}, and \textbf{Naive Coordinate}. All three use the same split-conformal calibration and therefore satisfy the marginal guarantee $\Prob(Y_{n+1}\in\calC(X_{n+1}))\ge 1-\alpha$. Their differences appear in the shape, size, and local behavior of the resulting regions. The base predictor is a $k$-NN regressor ($k\!=\!20$) followed by extrinsic mean projection onto~$\Sph$. The difficulty estimator $\shat(x)$ is trained by 5-fold cross-validation on the training set alone.

\begin{figure*}[t]
\centering
\includegraphics[width=\textwidth]{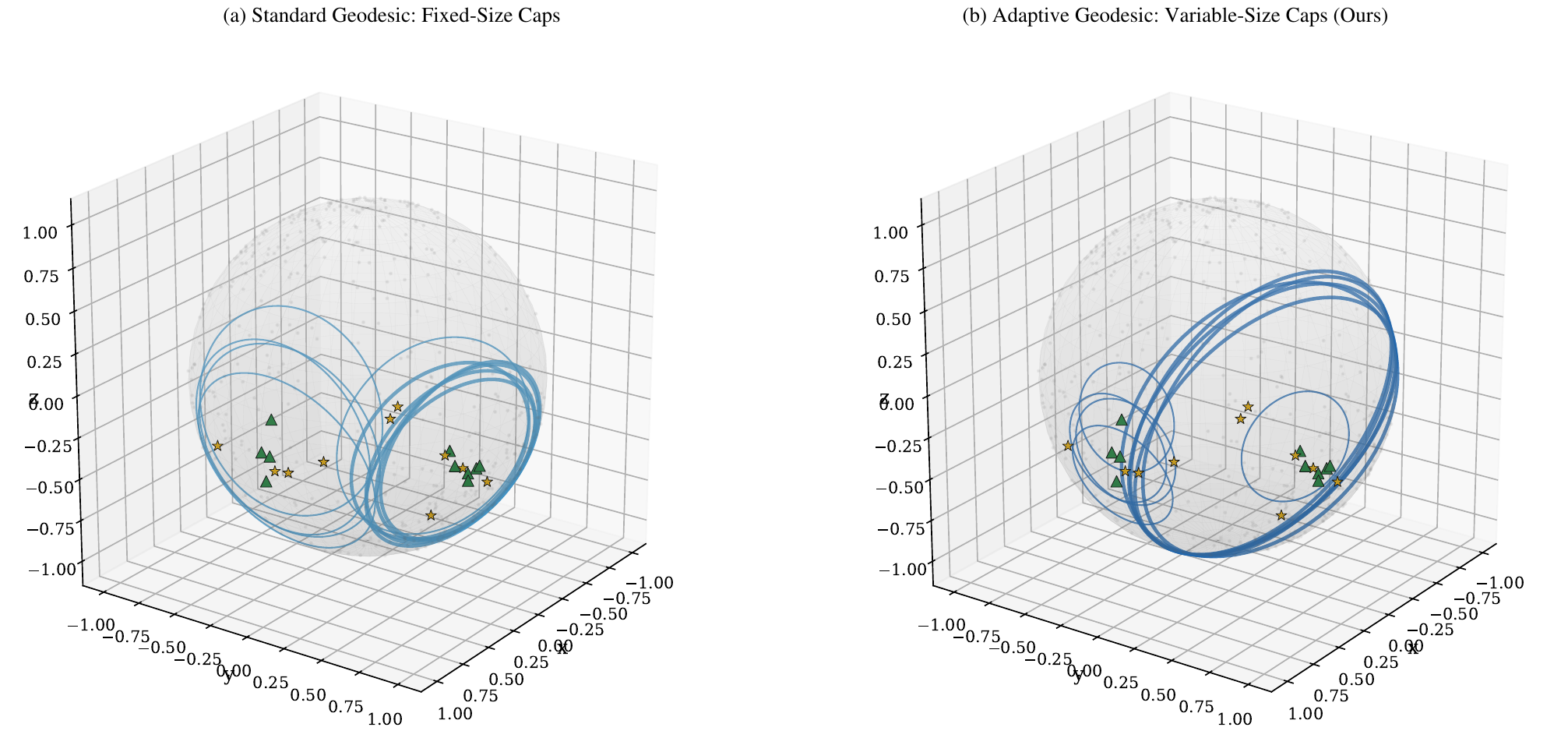}
\caption{Visual comparison of the two geodesic conformal methods on the synthetic $\Sph$ benchmark. \textbf{(a)}~The standard geodesic method assigns a fixed-radius cap to every test point, so easy and hard cases receive identical-size regions. \textbf{(b)}~The proposed adaptive method varies cap size by local difficulty $\shat(x)$, producing tighter regions where prediction is easier and wider regions where prediction is harder.}
\label{fig:syn_3d}
\end{figure*}

Figure~\ref{fig:syn_3d} shows the main qualitative effect on a slice of the synthetic benchmark. The standard geodesic method (panel~a) assigns the same cap size to every test point. The adaptive method (panel~b) uses smaller caps where the fitted predictor is locally accurate and larger caps where the task is harder.

\subsection{Case Study 1: Synthetic Sphere}

We sample $n = 1{,}200$ points with covariates $x \in [-3,3]^2$ and responses $Y \sim \mathrm{vMF}(\mu(x), \kappa(x))$. The concentration $\kappa(x) = 3 + 147\exp(-\|x\|^2/4)$ changes by a factor of $50$, from highly dispersed responses near the boundary ($\sim\!33^{\circ}$ error) to tightly concentrated responses near the center ($\sim\!4.7^{\circ}$ error). Table~\ref{tab:synthetic} reports 300-trial results at $\alpha = 0.10$. All methods achieve marginal coverage near the nominal $0.90$. The adaptive geodesic method reduces the standard deviation of conditional coverage by 19\% and raises worst-bin coverage from 0.814 to 0.839. The naive coordinate method needs 26\% more area than the geodesic methods, reflecting the cost of chart distortion.

\begin{table}[!htb]
\centering
\caption{Synthetic $\Sph$: 300 trials, $\alpha = 0.10$, $n = 1{,}200$, $50\times$ heteroscedasticity.}
\label{tab:synthetic}
\smallskip
\small
\begin{tabular}{@{}lcccc@{}}
\toprule
\textbf{Method} & \textbf{Marg.} & \textbf{Area} & \textbf{Cond.} & \textbf{Worst} \\
 & \textbf{Cov.} & \textbf{(sr)} & \textbf{Std$\downarrow$} & \textbf{Cov.$\uparrow$} \\
\midrule
Adaptive Geo. & $0.906$ & 1.865 & \textbf{0.042} & \textbf{0.839} \\
Standard Geo. & $0.904$ & 1.885 & 0.052 & 0.814 \\
Naive Coord.  & $0.904$ & 2.376 & 0.067 & 0.784 \\
\bottomrule
\end{tabular}
\end{table}

\begin{figure}[!htb]
\centering
\includegraphics[width=\linewidth]{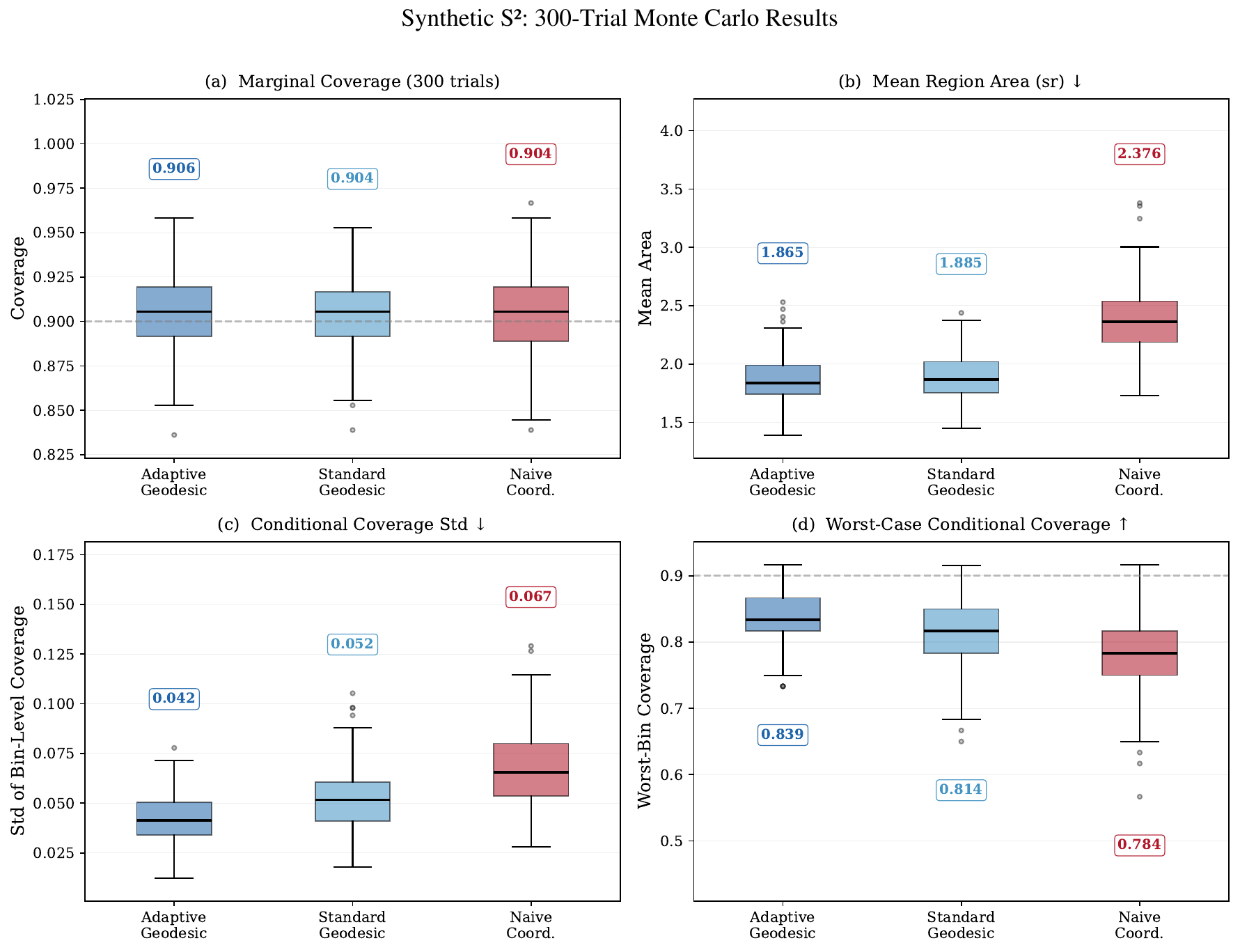}
\caption{300-trial Monte Carlo distributions on the synthetic $\Sph$ benchmark. \textbf{(a)}~All methods achieve valid marginal coverage near $0.90$. \textbf{(b)}~The naive coordinate method requires $\approx\!26\%$ more area. \textbf{(c)}~The adaptive method attains the lowest conditional coverage standard deviation. \textbf{(d)}~Adaptive worst-case coverage stays closest to the $0.90$ target.}
\label{fig:syn_trials}
\end{figure}

Figure~\ref{fig:syn_trials} shows the full Monte Carlo distributions. Marginal coverage is similar for all methods (panel~a), but the adaptive method is more stable across trials in area, conditional coverage, and worst-bin coverage (panels b to d).

\begin{figure}[tb]
\centering
\includegraphics[width=\linewidth]{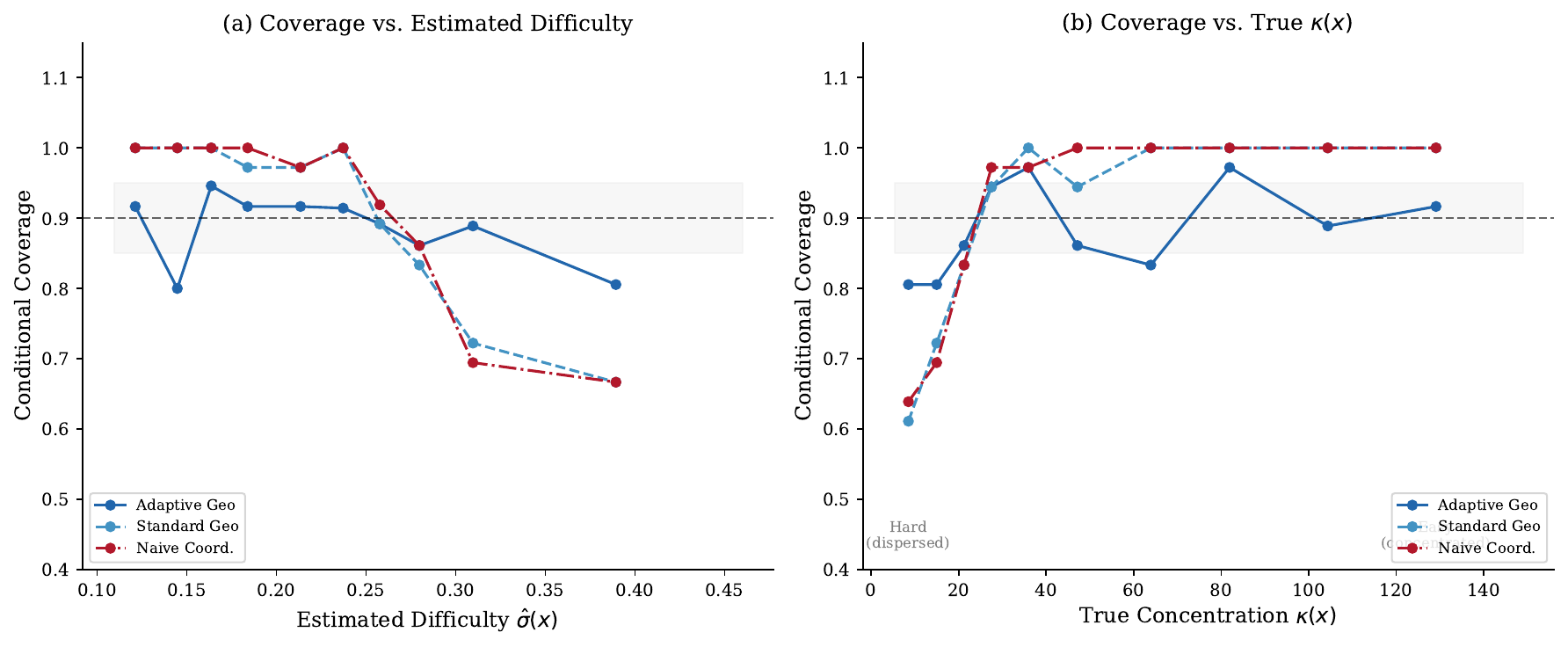}
\caption{Conditional coverage on the synthetic benchmark. \textbf{(a)}~Coverage vs.\ estimated difficulty $\shat(x)$: standard and naive methods over-cover easy bins ($>\!0.97$) and drop below $0.70$ on hard bins; the adaptive method stays close to $0.90$ across the range. \textbf{(b)}~Coverage vs.\ true vMF concentration $\kappa(x)$: at low~$\kappa$ (hard, dispersed targets), both non-adaptive methods fall below $0.90$, while the adaptive method remains comparatively stable.}
\label{fig:syn_conditional}
\end{figure}

Figure~\ref{fig:syn_conditional} explains where the gain comes from. When test points are binned by estimated difficulty $\shat(x)$ (panel~a), the two non-adaptive methods over-cover easy regions and under-cover hard regions. The adaptive method stays much closer to the $0.90$ target across the range. Binning by the true vMF concentration $\kappa(x)$ (panel~b) gives the same message using ground-truth difficulty: the non-adaptive methods lose coverage on dispersed targets, while the adaptive method remains comparatively flat.

\subsection{Case Study 2: IGRF-14 Geomagnetic Forecasting}

We next build a temporal forecasting task from the International Geomagnetic Reference Field, using the IGRF-14 coefficients distributed by NOAA/NCEI together with the standard IGRF citation \citep{alken2021igrf,noaa2024igrf14}. At 352 surface locations, consisting of 52 INTERMAGNET observatories and 300 random sites, we compute the geomagnetic field unit vector $\hat{\mathbf{B}}(x,t) \in \Sph$ at 38 semi-annual epochs from 2005 to 2023. The task is to predict $\hat{\mathbf{B}}(x, t{+}1\,\text{yr})$ from $(\text{lat}, \text{lon}, \hat{\mathbf{B}}(x,t), t)$, producing 3,000 subsampled forecast pairs.

\begin{figure*}[t]
\centering
\includegraphics[width=\textwidth]{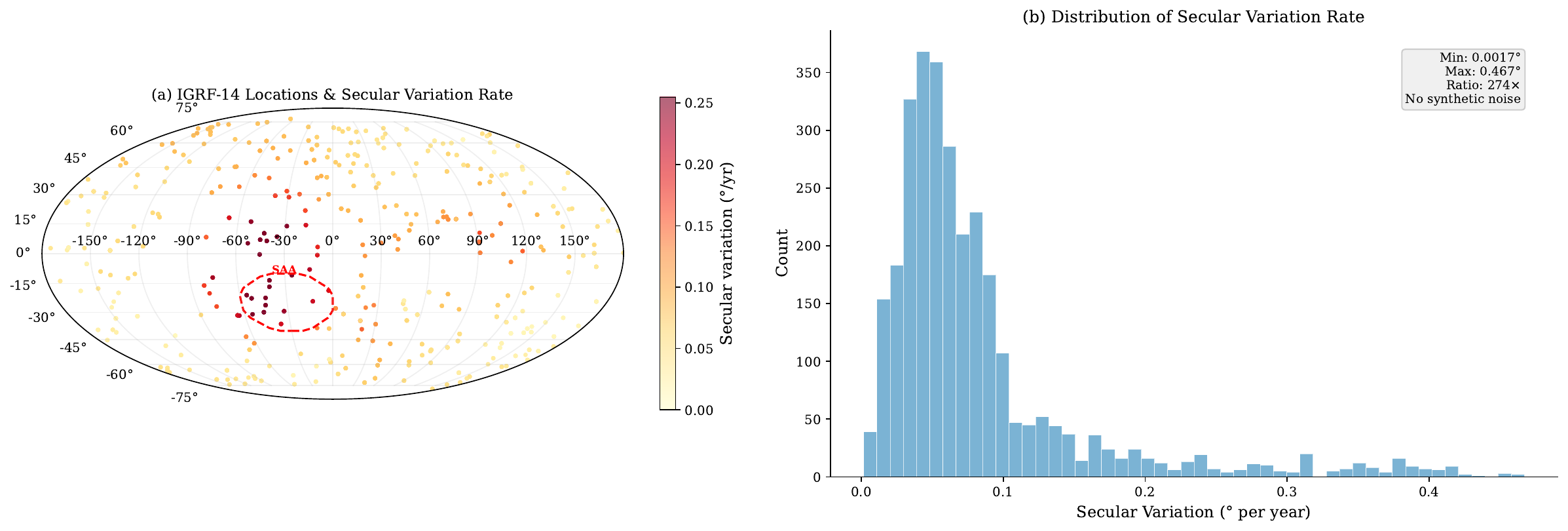}
\caption{IGRF-14 data overview. \textbf{(a)}~Secular variation rate ($^{\circ}$/yr) at the 352 surface sites. The South Atlantic Anomaly (SAA, dashed ellipse) and magnetic equator exhibit the fastest field evolution, creating intrinsic spatial heteroscedasticity. \textbf{(b)}~Distribution of the secular variation rate, spanning a $274\times$ range from $0.0017^{\circ}$/yr to $0.467^{\circ}$/yr. No synthetic noise is added.}
\label{fig:geo_data}
\end{figure*}

No synthetic noise is added. The residuals come from the $k$-NN model's imperfect extrapolation of real secular variation, which is naturally heteroscedastic over the globe. Figure~\ref{fig:geo_data} shows this structure: the South Atlantic Anomaly and the magnetic equator (panel~a) have the fastest field evolution, while polar regions are nearly stationary. The fastest and slowest sites differ by a factor of $274$ (panel~b), making this a useful test of local adaptivity.

\begin{table}[!htb]
\centering
\caption{IGRF-14 geomagnetic forecasting ($\Sph$): 100 trials, $\alpha = 0.10$, $n = 3{,}000$. Difficulty estimator: $r(\shat, \text{residual}) = 0.516$.}
\label{tab:igrf}
\smallskip
\small
\begin{tabular}{@{}lcccc@{}}
\toprule
\textbf{Method} & \textbf{Marg.} & \textbf{Area} & \textbf{Cond.} & \textbf{Worst} \\
 & \textbf{Cov.} & \textbf{(sr)} & \textbf{Std$\downarrow$} & \textbf{Cov.$\uparrow$} \\
\midrule
Adaptive Geo. & $0.902$ & 0.038 & \textbf{0.031} & \textbf{0.855} \\
Standard Geo. & $0.902$ & 0.039 & 0.107 & 0.689 \\
Naive Coord.  & $0.903$ & 0.046 & 0.060 & 0.805 \\
\midrule
\multicolumn{3}{@{}l}{\footnotesize\textit{Wilcoxon (Adapt.\ vs.\ Std.):}} & \footnotesize $p{<}4{\times}10^{-18}$ & \footnotesize $p{<}4{\times}10^{-18}$ \\
\bottomrule
\end{tabular}
\end{table}

\begin{figure}[!htb]
\centering
\includegraphics[width=\linewidth]{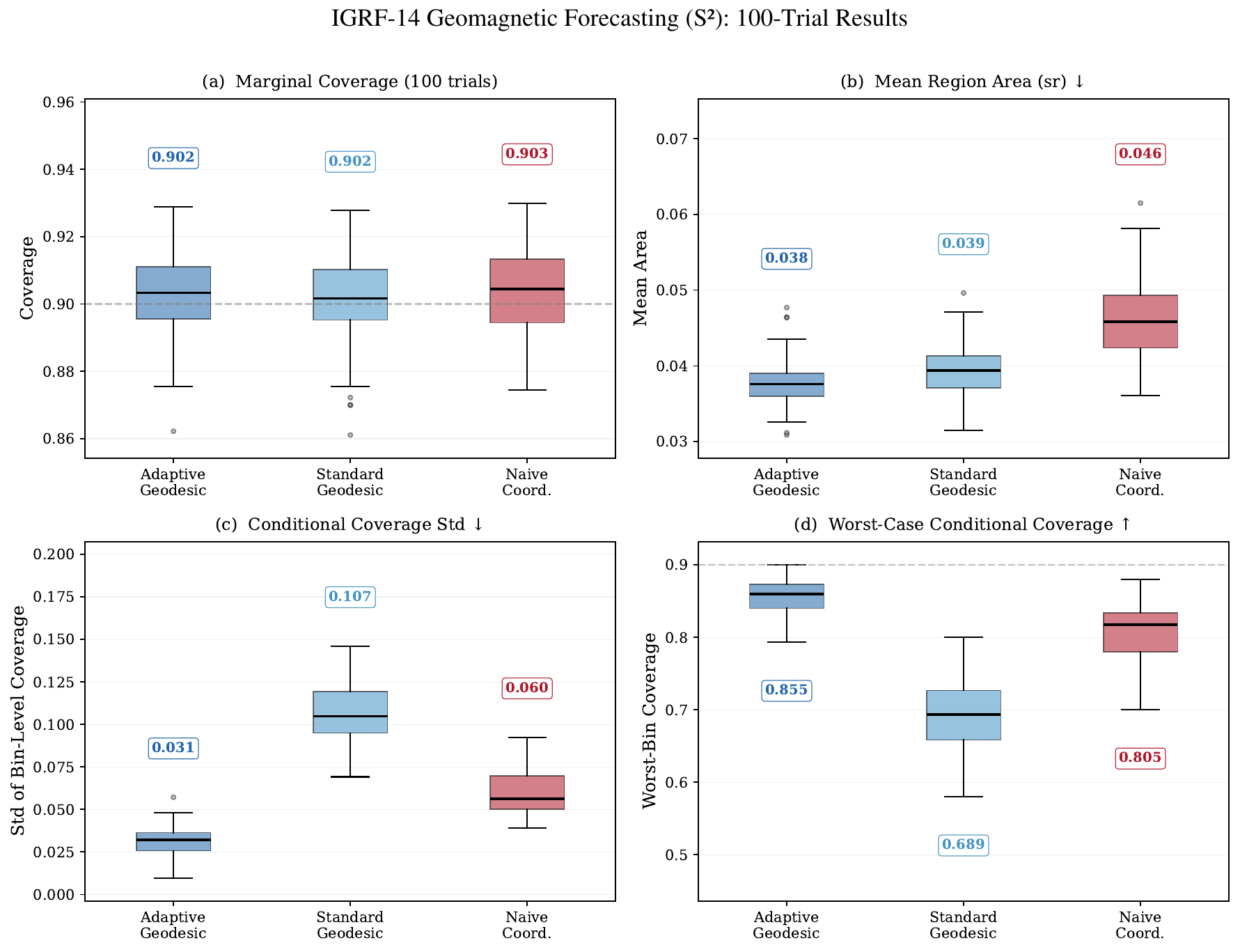}
\caption{IGRF-14 100-trial Monte Carlo distributions. \textbf{(a)}~All methods achieve valid marginal coverage near $0.90$. \textbf{(b)}~The naive coordinate method uses substantially more area. \textbf{(c)}~The adaptive method reduces conditional coverage standard deviation by $\sim\!3.5\times$ relative to standard geodesic. \textbf{(d)}~Worst-case bin coverage improves by $0.166$ relative to standard geodesic.}
\label{fig:geo_trials}
\end{figure}

Table~\ref{tab:igrf} and Figure~\ref{fig:geo_trials} summarize the 100-trial results. The difficulty estimator has Pearson correlation $r = 0.516$ ($p < 10^{-123}$) with actual residuals, indicating that $\shat(x)$ captures a meaningful part of the spatial variation in error. The adaptive method reduces the conditional coverage standard deviation by 71\% (0.031 vs.\ 0.107) and raises worst-bin coverage from 0.689 to 0.855. These differences are highly significant under a paired Wilcoxon signed-rank test.

\begin{figure*}[t]
\centering
\includegraphics[width=\textwidth]{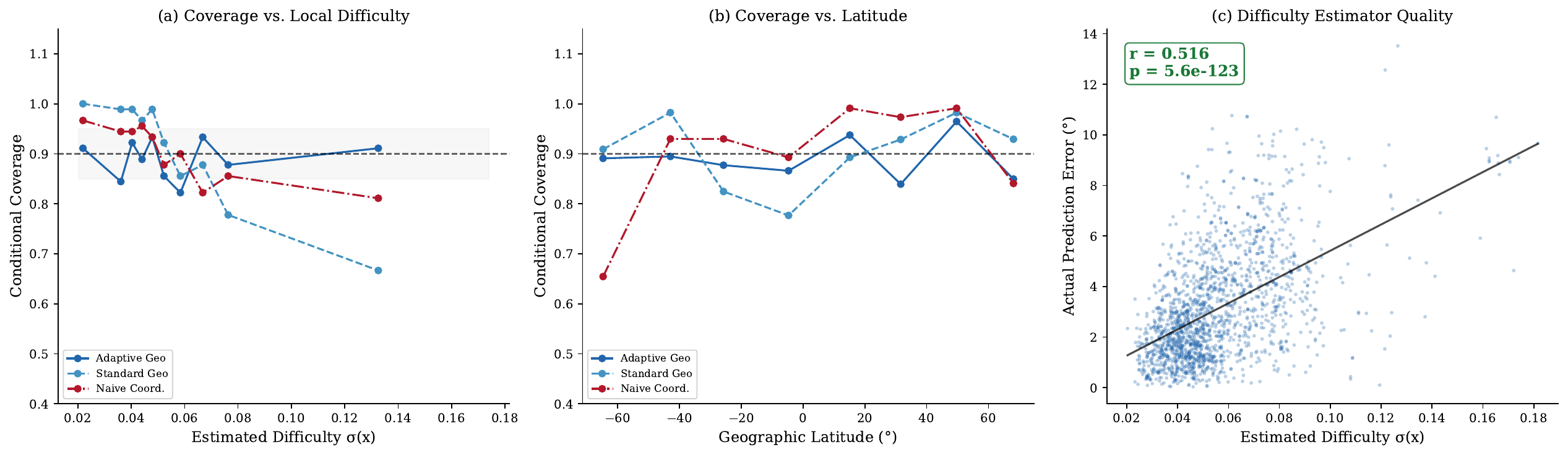}
\caption{IGRF-14 conditional coverage and diagnostic. \textbf{(a)}~Coverage vs.\ estimated difficulty $\shat(x)$: the standard geodesic method drops below $0.70$ in the hardest bin, while the adaptive method stays near $0.90$. \textbf{(b)}~Coverage vs.\ geographic latitude: the standard method under-covers equatorial latitudes where secular variation is fastest. \textbf{(c)}~Difficulty estimator quality: $\shat(x)$ vs.\ actual prediction error ($r = 0.516$, $p \approx 5.6 \times 10^{-123}$).}
\label{fig:geo_conditional}
\end{figure*}

Figure~\ref{fig:geo_conditional} shows the geographic structure of the improvement. The standard geodesic method under-covers equatorial regions (panel~b), where secular variation is fastest and reliable uncertainty quantification is especially important. The adaptive method keeps coverage closer to nominal across both latitude (panel~b) and estimated difficulty (panel~a). The correlation between estimated difficulty and actual error (panel~c) explains why the normalization is effective.

\subsection{Discussion}

Across both experiments, the adaptive geodesic method gives the most uniform conditional coverage and the strongest worst-bin coverage while preserving the marginal guarantee. The improvement comes from two complementary choices. The first is adaptivity: scaling by $\shat(x)$ makes conformity scores more comparable across easy and hard regions. The second is geometry: geodesic caps have position-independent area on $\Sph$, whereas coordinate rectangles spend area unevenly because of metric distortion. The correlation between $\shat(x)$ and the residual can be used as a practical diagnostic. In the IGRF-14 experiment, $r = 0.516$ is associated with a 71\% reduction in conditional coverage standard deviation. If this correlation is weak on a validation set, for example below about $0.15$, the standard geodesic method may be preferable.

\section{Conclusion}

We introduced adaptive geodesic conformal prediction for regression with manifold-valued responses. The method uses intrinsic distances to respect the geometry of the response space and a cross-validated difficulty estimator to adjust region size across the input space. On a synthetic sphere task and a geomagnetic forecasting task, this combination improves conditional coverage uniformity and worst-case coverage while preserving valid marginal coverage. A natural next step is to extend the framework to streaming or otherwise non-exchangeable data. Another useful direction is to move beyond isotropic geodesic caps toward anisotropic regions that can represent direction-dependent uncertainty on the manifold.

\FloatBarrier
{\small
\setlength{\bibsep}{-1pt}
\bibliography{references}
\bibliographystyle{plainnat}
}

\end{document}